\documentclass[11pt]{article}
\usepackage[utf8]{inputenc}

\usepackage[margin=1in]{geometry}
\usepackage{hyperref,setspace,tabularx,booktabs,amsbsy,amsmath,amsfonts,amsthm,bm, amssymb,authblk,xcolor,colortbl,multirow,graphicx, nicefrac,float,caption,subcaption,comment,fancyhdr,natbib,bibunits, threeparttable}
\usepackage{bbm}
\usepackage{longtable}% for long tables
\usepackage{multirow}
\usepackage{bm}
\usepackage{booktabs}
\usepackage{bm}
\usepackage{mathtools}
\usepackage{amsmath}
\usepackage{amssymb}
\usepackage{algorithm}
\usepackage{algorithmic}
\usepackage{amsthm}
\usepackage{lscape}
\usepackage{xcolor}

\theoremstyle{plain}

\theoremstyle{definition}

\theoremstyle{remark}

\DeclarePairedDelimiterX{\infdivx}[2]{(}{)}{%
  #1\;\delimsize\|\;#2%
}

\newcommand{\infdivkl}{D_{KL}\infdivx}

\DeclareMathOperator{\pa}{\textbf{PA}}

\title{Bayesian Meta-Learning for Improving Generalizability of Health Prediction Models With Similar Causal Mechanisms}

\date{\vspace{-5ex}}

\author[1]{Sophie Wharrie}
\author[2]{Lisa Eick}
\author[1]{Lotta Mäkinen}
\author[2,3]{Andrea Ganna}
\author[1,4]{Samuel Kaski}
\author[1]{FinnGen}

\affil[1]{Department of Computer Science, Aalto University, Espoo, Finland}
\affil[2]{Institute for Molecular Medicine Finland, Helsinki Institute of Life Science, University of Helsinki, Helsinki, Finland}
\affil[3]{Massachusetts General Hospital and Broad Institute of MIT and Harvard, Cambridge, MA, USA}
\affil[4]{Department of Computer Science, University of Manchester, United Kingdom}

\begin{document}
    \maketitle

    \section*{Abstract} 

Machine learning strategies like multi-task learning, meta-learning, and transfer learning enable efficient adaptation of machine learning models to specific applications in healthcare, such as prediction of various diseases, by leveraging generalizable knowledge across large datasets and multiple domains. In particular, Bayesian meta-learning methods pool data across related prediction tasks to learn prior distributions for model parameters, which are then used to derive models for specific tasks. However, inter- and intra-task variability due to disease heterogeneity and other patient-level differences pose challenges of negative transfer during shared learning and poor generalizability to new patients. We introduce a novel Bayesian meta-learning approach that aims to address this in two key settings: (1) predictions for new patients (same population as the training set) and (2) adapting to new patient populations. Our main contribution is in modeling similarity between causal mechanisms of the tasks, for (1) mitigating negative transfer during training and (2) fine-tuning that pools information from tasks that are expected to aid generalizability. We propose an algorithm for implementing this approach for Bayesian deep learning, and apply it to a case study for stroke prediction tasks using electronic health record data. Experiments for the UK Biobank dataset as the training population demonstrated significant generalizability improvements compared to standard meta-learning, non-causal task similarity measures, and local baselines (separate models for each task). This was assessed for a variety of tasks that considered both new patients from the training population (UK Biobank) and a new population (FinnGen).

\section{Introduction}
\label{sec:introduction}

Given a dataset of related tasks, how can we train a machine learning model that performs well on new data? In this work we study this fundamental problem of generalization in machine learning for a setting of related supervised learning tasks (i.e., learning to predict outputs $\bm{y}_t$ from inputs $\bm{x}_t$ for tasks $t$). Motivated by applications of machine learning for personalized health with electronic health record (EHR) and biobank datasets, we focus on a realistic setting where data is not independent and identically distributed (i.i.d.), due to both inter- and intra-task variability. Owing to the scale, breadth and longitudinal nature of these rich data sources, machine learning is increasingly being used for applications including disease prediction, patient trajectory modeling, and clinical decision support \cite{shickel2017deep}. However, for complex diseases such as type 2 diabetes, Alzheimer's disease and stroke, data variability arises from different subtypes of disease and individual-level differences in patient outcomes driven by different underlying disease mechanisms and treatment responses \cite{nair2022heterogeneity, scheltens2016identification}. 

Machine learning in these non-i.i.d. settings can be formulated in terms of learning from multiple related tasks originating from a common task environment, where some aspects of the data are common across all tasks (global properties), but other aspects are unique to a specific task (local properties). For example, the different tasks may represent subtypes of a disease or even individual patients. The general problem has been examined in learning paradigms such as meta-learning (or ``learning how to learn'') and multi-task learning \cite{caruana1997multitask, hospedales2021meta}, which have been successfully applied to non-i.i.d. datasets in healthcare and related domains \cite{rafiei2023meta}. Promising applications include predicting disease outcomes and survival analysis in limited data settings, for example, to improve predictions for rare diseases by leveraging relevant information from more common diseases \cite{zhang2019metapred, tan2022metacare, qiu2020meta}. 

Since this work concerns generalizability challenges, we focus on Bayesian meta-learning methods \cite{hospedales2021meta}, which allow for studying two settings of interest: (1) predictions for new patients (sampled from the training population), and (2) adapting to new patient populations. However, standard meta-learning strategies do not explicitly address several generalizability challenges arising from inter- and intra- task variability. First, negative transfer may occur when the tasks being learned together are not sufficiently similar, leading to degraded performance compared to single-task learning \cite{wuunderstanding, pmlr-v119-standley20a}. General solutions to this problem include identifying the subset of tasks for training that are sufficiently related to the target domain, or dynamically adjusting the influence of each task during training \cite{yu2020gradient, shui2019principled, liu2019loss}. While these approaches have been extensively studied in the computer vision and natural language processing domains, the development of specialized methods for the health domain remains less explored. Second, machine learning models may struggle to generalize to new patients underrepresented during training. A known problem for machine learning with EHR data is the phenomenon of ``shortcut learning'', where ML models rely on spurious associations in the training data \cite{brown2023detecting, geirhos2020shortcut}. The overreliance on shortcuts becomes problematic when they do not generalize to new patients or other changes between the source and target domains. General strategies motivated by causal inference techniques have been proposed to mitigate these issues, for example, by learning from invariant causal mechanisms in the data that persist across the source and target domains \cite{peters2016causal, rojas2018invariant, magliacane2018domain}. However, to the best of our knowledge, the practical implementation of these approaches has not been explored in meta-learning for health prediction tasks.

As a motivating example, in a setting where machine learning is applied to various disease prediction tasks, certain diseases would be more relevant for improving the prediction of a specific disease. However, even within one disease prediction task, there may be significant variability among patients that was not completely observed in the training data because of small sample sizes. Therefore, the machine learning algorithm needs to not only identify the most relevant tasks (addressing inter-task variability) and learn from them, but also adapt to new patients (addressing intra-task variability).

Bayesian meta-learning, formalized in terms of hierarchical Bayesian modeling, provides a principled machine learning framework for studying how statistical strength can be shared across related prediction tasks to aid generalizability \cite{grant2018recasting, gordon2019meta, ravi2018amortized, amit2018meta}. Specifically, Bayesian meta-learning methods pool data across multiple tasks to learn prior distributions for model parameters, which are then used with additional data to derive posterior distributions for specific task models. Previous works have explored how modeling task similarity in (non-Bayesian) meta-learning can improve predictive performance \cite{zhou2021task, yao2019hierarchically, yao2020automated} (i.e., measuring similarity in the feature, model parameter, or meta-knowledge space). In the health domain, differences in the tasks arise from different mechanisms of disease or treatment responses. However, these causal mechanisms are typically unknown and infeasible to establish from observational data alone. Therefore, standard statistical methods for identifying similar tasks may not necessarily find tasks that aid generalizability (e.g., if the task similarity measure captures spurious associations in patient demographics instead of the underlying biology). Our main contribution is a novel Bayesian meta-learning approach that improves generalizability by modeling similarity between causal mechanisms of the tasks. This is incorporated during meta-training to mitigate negative transfer and also at the fine-tuning stage to pool information from tasks that are expected to aid generalizability to new populations.

Section \ref{sec:methods} of this paper introduces our Bayesian meta-learning approach for learning from multiple tasks in non-i.i.d. settings. We establish baselines of separate models for each task and a standard Bayesian meta-learning approach. Formalized as hierarchical Bayesian modeling, we then extend the standard Bayesian meta-learning approach to focus on learning from related tasks. We compare several measures for identifying similar tasks: three approaches motivated by causal inference techniques (Mendelian Randomization \cite{sanderson2022mendelian}, structure learning \cite{lorch2021dibs}, invariant causal prediction \cite{peters2016causal}), and one baseline (independence testing).

Section \ref{sec:case_study} presents results from a case study using the UK Biobank and FinnGen datasets. Meta-learning is applied to binary classification tasks for predicting stroke and related diseases. Generalizability is assessed by meta-training on the UK Biobank dataset and evaluating predictive performance by adapting (fine-tuning) to target tasks of interest (stroke prediction) for the same (UK Biobank) and new (FinnGen) patient populations. We compare baseline models with the proposed task similarity-based meta-learning models in terms of their ability to mitigate negative transfer and improve generalizability, further supported by analyses of feature importance, patient characteristics and task similarity estimates. Section \ref{sec:conclusion} discusses the implications of our findings and suggests directions for future research.

\section{Methods}
\label{sec:methods}

\begin{figure*}[!t]
\centerline{\includegraphics[width=0.8\linewidth]{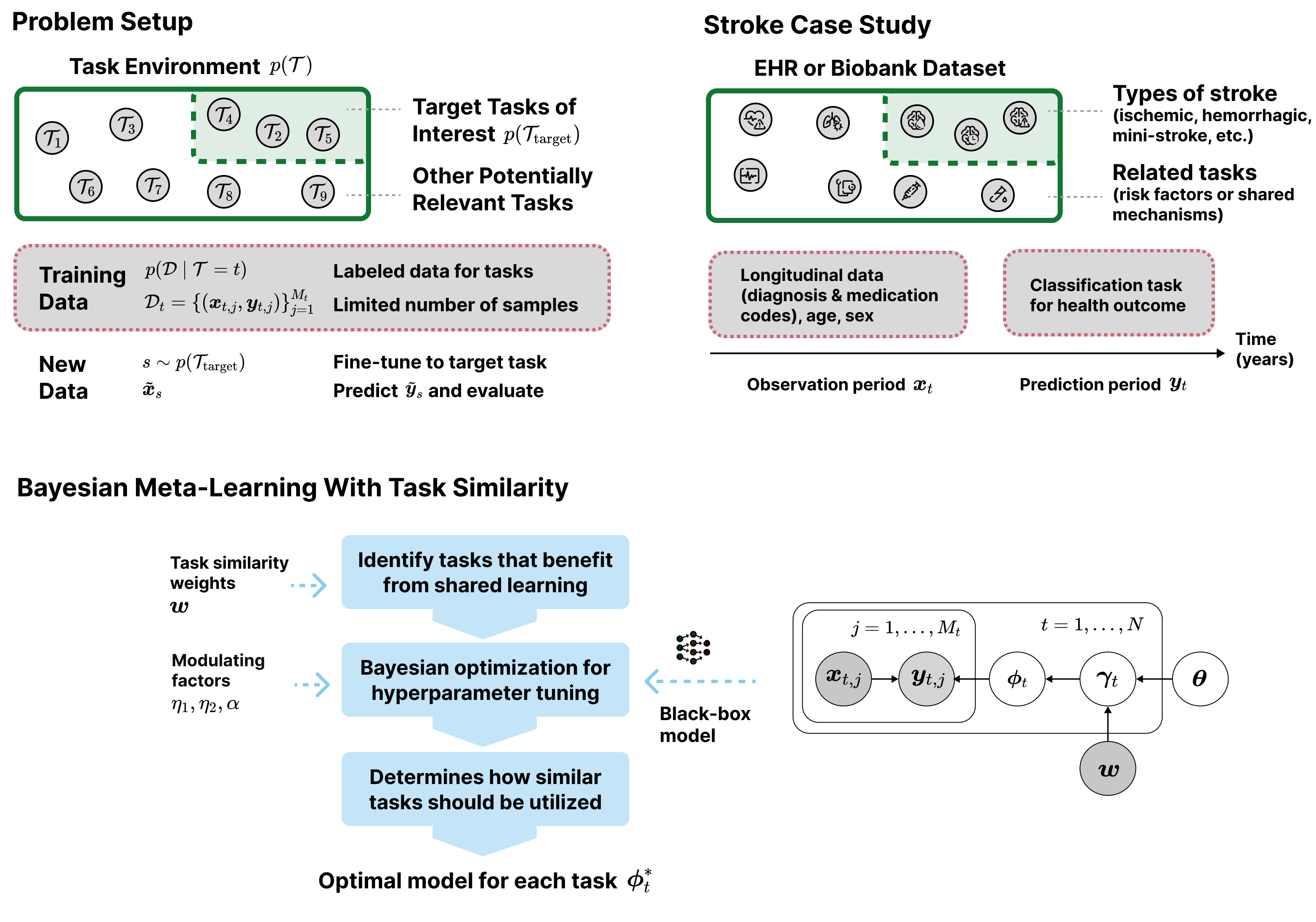}}
\caption{Problem setup and methods. Top left: General problem setup for task environment and data. Top right: Specific case study for stroke prediction. Bottom: Bayesian meta-learning with task similarity - a hierarchical model is used for deriving task-specific models ($\bm{\phi}_t$) from global ($\bm{\theta}$) parameters shared across all tasks and parameters ($\bm{\gamma}_t$) shared across related tasks, governed by task similarity weights ($\bm{w}$).}
\label{fig:graphicalabstract}
\end{figure*}

\subsection{Problem Setup}

The problem setup studied in this work is formulated as follows and illustrated in Figure \ref{fig:graphicalabstract}. Each task $t$ is sampled from a task environment defined by a task distribution $t \sim p(\mathcal{T})$ (standard setup in studying inductive bias, \cite{baxter2000model}). A task has $M_t$ labeled samples $\mathcal{D}_t = \{ (\bm{x}_{t,j}, \bm{y}_{t,j}) \}_{j=1}^{M_t}$ from a task-specific data distribution $p(\mathcal{D} \mid \mathcal{T}=t)$, where $\bm{x}_{t,j}\in \mathcal{X} \subseteq \mathbb{R}^K$ consists of the same $K$ variables for all tasks and the goal is to learn how to predict outputs $\bm{y}_t$ from inputs $\bm{x}_t$. During meta-training, a model is trained for $N$ tasks sampled from the task environment $t \sim p(\mathcal{T})$, and at evaluation (or fine-tuning) time, labeled data is available for a subset of target tasks of interest $s \sim p(\mathcal{T}_{\text{target}})$. This corresponds to a setting where the goal is to ``learn how to learn'' machine learning models for the target tasks (e.g., stroke subtypes), also utilizing a broader set of potentially relevant tasks from EHRs. To study two generalizability settings of interest, first we assume that all tasks come from the same patient population, and later consider fine-tuning on patients from a different population. 

\subsection{Learning From Similar Tasks}

We use a hierarchical Bayesian model (HBM) as the basis of the Bayesian meta-learning method for this work. A basic HBM consists of global parameters $\bm{\theta}$ that act to pool information across multiple tasks and task-specific parameters $\bm{\phi}_t$ that allow for variation between tasks. HBMs therefore provide a general probabilistic framework for formalizing the key ideas of multi-task learning, meta-learning and related strategies \cite{baxter2000model, grant2018recasting, gordon2019meta}. We are specifically interested in models that generalize well to unseen data samples for the target tasks. However, a key challenge for health data is limited sample sizes for training tasks that do not fully capture patient-level variation that could be encountered in the target tasks. Meta-learning aims to address this by ``learning to learn'' machine learning models that generalize across a set of tasks and to new tasks \cite{hospedales2021meta, murphy2023probabilistic}. In standard Bayesian meta-learning, formulated as a HBM, the global model can be viewed as a prior distribution for the parameters of task-specific models, with information sharing via hierarchical priors $p(\bm{\phi}_t \mid \bm{\theta}), p(\bm{\theta})$ \cite{grant2018recasting, gordon2019meta, fortuin2022priors}. It follows that the posterior predictive distribution for a task, given the observed data $ \mathcal{D}_t$, is computed as

% The task data $\mathcal{D}_t$ for training is split into support $\mathcal{D}_t^{(s)}$ and query $\mathcal{D}_t^{(q)}$ sets. The support set is used to adapt the base-learner to the task at hand, and the query set is used to update the meta-learner's ability to generalize to new data from the same task. 

\begin{equation}
p(\tilde{\bm{y}}_t \mid \tilde{\bm{x}}_t, \bm{\theta}) = \int p(\tilde{\bm{y}}_t \mid \tilde{\bm{x}}_t, \bm{\phi}_t) p(\bm{\phi}_t \mid \mathcal{D}_t, \bm{\theta}) d \bm{\phi}_t.
\label{eq:hbmpredposterior}
\end{equation}

For health data, it may not be the case that all tasks are as equally important for improving learning of a specific task - for example, negative transfer can occur when the tasks are not sufficiently similar \cite{wuunderstanding, pmlr-v119-standley20a}.  Prior works have extended (non-Bayesian) meta-learning algorithms to consider task similarity (e.g., similarity in the feature space, model parameter space, or meta-knowledge space \cite{zhou2021task, yao2019hierarchically, yao2020automated}). A key contribution of our work is formalizing task similarity within the Bayesian meta-learning setting and exploring forms of task similarity relevant for health data. Figure \ref{fig:graphicalabstract} illustrates how we extend the Bayesian meta-learning model with hierarchical priors that assume a task-specific model $\bm{\phi}_t$ is derived from a global model shared across all tasks $\bm{\theta}$ and parameters $\bm{\gamma}_t$ shared across the most related tasks (determined by a matrix of inter-task similarity weights $\bm{w}$). The posterior distribution over the model parameters becomes

\begin{equation}
\begin{aligned}
p(&\bm{\theta}, \bm{\gamma}_1, ..., \bm{\gamma}_N, \bm{\phi}_1, ..., \bm{\phi}_N \mid \bm{w}, \mathcal{D}_1, ..., \mathcal{D}_N) = \\
&p(\bm{\theta} \mid \mathcal{D}_1, ..., \mathcal{D}_N) \prod_{t=1}^N p(\bm{\gamma}_t \mid \bm{\theta}, \bm{w}, \mathcal{D}_1, ..., \mathcal{D}_N) p(\bm{\phi}_t \mid \bm{\gamma}_t, \mathcal{D}_t),
\end{aligned}
\label{eq:hbmjointposterior}
\end{equation}

In Section \ref{ssec:bnnmethods} we describe algorithms implementing this using variational inference techniques for Bayesian neural networks used in experiments.

\subsection{Identifying Similar Tasks}
\label{ssec:tasksimmethods}

In EHR datasets, where a large number of candidate tasks (e.g., diagnosis and medication labels) can be relevant to target tasks of interest, it is beneficial to utilize prior knowledge encoded as a task similarity matrix $\bm{w}$ to inform task selection. For calculating these weights, task similarity can be defined in terms of similarity in the tasks' joint distributions $p(\bm{x}_t, \bm{y}_t)$ or predictive distributions $p(\bm{y}_t \mid \bm{x}_t)$. However, distributions inferred from observational health data are known to learn from spurious associations - a phenomenon known as ``shortcut learning'' \cite{geirhos2020shortcut}. For instance, machine learning models have been observed to exploit correlations with sensitive attributes (age, sex, race, etc.) or non-clinically relevant features (such as the type of imaging equipment) \cite{brown2023detecting, zech2018variable}. By comparing several approaches for computing task similarity in this work (including causal inference-based techniques), we are interested in understanding whether the estimates reflect clinically or biologically meaningful ideas of task similarity and the implications for generalizability.

To reason about \textit{causal} task similarity, we theoretically formalize the task distributions in terms of causal models using structural causal models (SCMs) \cite{pearl2009causality}. A task SCM is described by a collection of structural equations $V_{t,j} := f_{t,j}(\pa_{t,j}^{\mathcal{G}_t}, N_{t,j})$ for $j=1,...,K+1$, where $\pa_{t,j}^{\mathcal{G}_t}$ are parent vertices of $V_{t,j}$ in the directed acyclic graph (DAG) $\mathcal{G}_t = (\bm{V}, \bm{E})$ consisting of $K+1$ vertices $\bm{V}=\{X_1, ..., X_K, Y\}$ and the edge set $\bm{E} \subseteq \bm{V}^2$, and $N_{t,j}$ is exogenous noise. The Markovian factorization expresses the joint distribution over the data $(X_1, ..., X_K, Y)$ as

\begin{equation}
\begin{aligned}
p(\mathcal{D} \mid \mathcal{T}=t) = \prod_{j=1}^{K+1} p(V_{t,j} \mid \pa_{t,j}^{\mathcal{G}_{t}}).
\end{aligned}
\label{eq:markovscm}
\end{equation}

\begin{figure*}[!t]
\centerline{\includegraphics[width=0.7\linewidth]{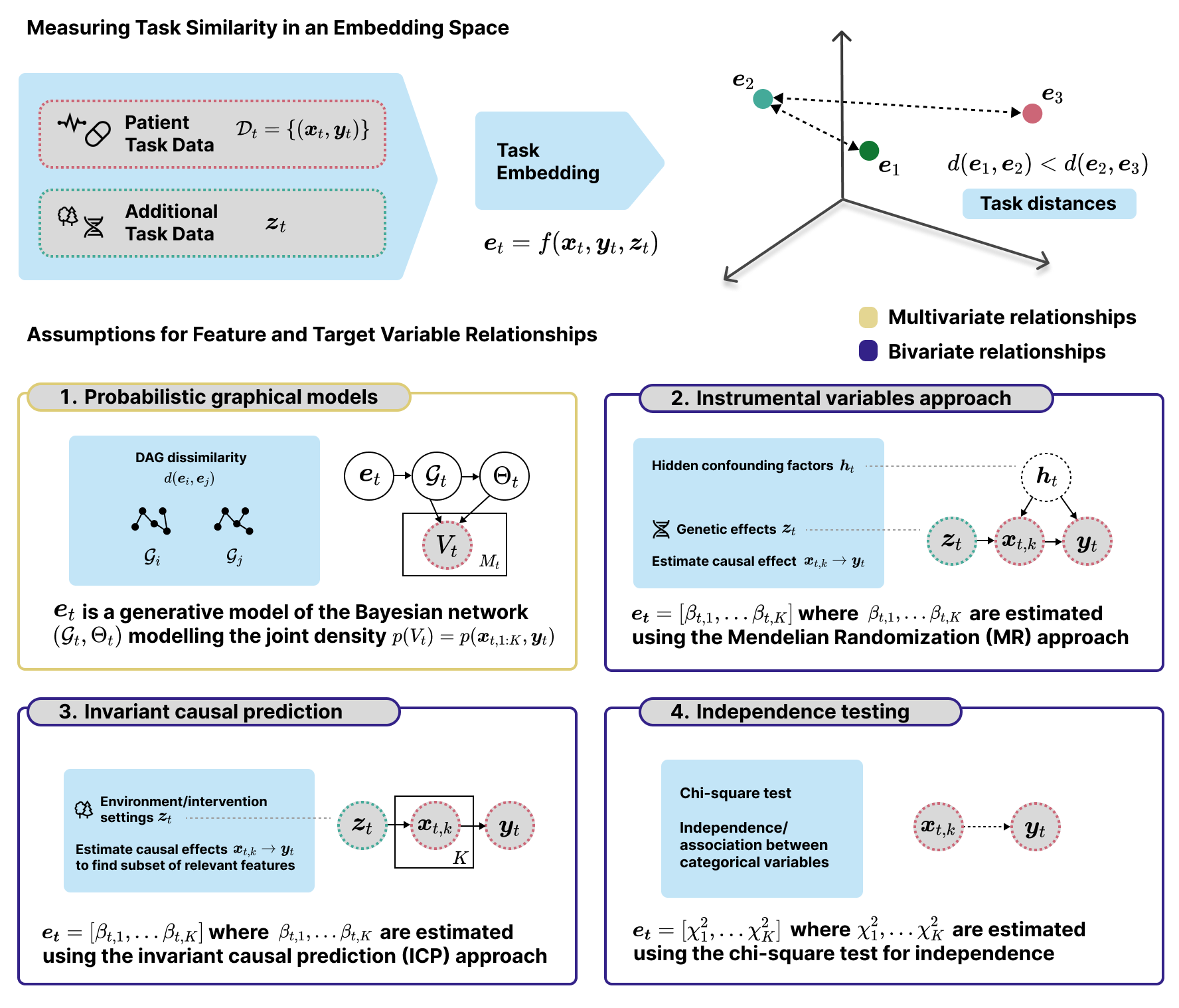}}
\caption{Task similarity is measured by mapping tasks to a space where similar tasks have small distances. Four methods are compared with different assumptions about the relationships between features, target variables, potential confounders, and additional variables that are assumed to aid identifiability of causal relationships: (1) Similarity in probabilistic graphical models, (2) Instrumental variables approach with Mendelian Randomization, (3) Invariant causal prediction across environments, and (4) Simple independence testing using the chi-square test.}
\label{fig:tasksimilaritydiagram}
\end{figure*}

Ideally, causal task similarity would be quantified in terms of similar SCMs, but in practice for EHR data, theoretical guarantees cannot be provided due to identification challenges in fitting SCMs to each task's limited observational data. Our implementation (Figure \ref{fig:tasksimilaritydiagram}) takes an approximate approach by embedding tasks $\mathcal{D}_t = \{(\bm{x}_t, \bm{y}_t)\}$ in a vector space $\bm{e}_t = f(\bm{x}_t, \bm{y}_t, \bm{z}_t)$, where small distances between embeddings are expected to indicate similar tasks. The addition of metadata $\bm{z}_t$ aims to facilitate the identification of causal relationships. We compare several approaches with different assumptions, chosen based on availability of suitable data:

\begin{enumerate}
    \item Similarity in DAG structures: We use the DiBS algorithm \cite{lorch2021dibs} to learn embeddings $\bm{e}_t$ that represent a generative model for the adjacency matrix $A_t$ of the directed graph $\mathcal{G}_t$ inferred from the task data. That is, $\bm{e}_t = \bm{u}_{t,i}^T \bm{v}_{t,j}$ and

\begin{equation}
\begin{aligned}
A_{t, ij} := \begin{cases}
\sigma_{\alpha}(\bm{u}_{t,i}^T \bm{v}_{t,j}) & \text{if } i \neq j \\
0 & \text{if } i = j,
\end{cases}
\end{aligned}
\label{eq:dibs}
\end{equation}

where $\sigma_{\alpha}$ is the sigmoid function with inverse temperature $\alpha$. This approach defines similar tasks in terms of similar DAGs. The DAGs model relationships between variables but the assumptions necessary for interpreting these as causal relationships, such as the absence of hidden confounders, are often untestable and/or violated in practice.
    
    \item Similarity using Mendelian Randomization (MR): MR utilizes genetic variants as instrumental variables to estimate the causal effect of an exposure on an outcome, even in the presence of hidden confounding \cite{sanderson2022mendelian}. Additional metadata $\bm{z}_t$ from genome-wide association studies (GWAS) is used to select instruments that make causal effects identifiable under the assumptions they are a) associated with the exposure, b) influence the outcome only through the exposure (i.e., no direct effect), and c) are independent of any hidden confounders affecting the exposure-outcome relationship. We use the two sample MR (2SMR) algorithm \cite{twosamplemr, mrsteiger} to construct $\bm{e}_t = [\beta_{1,t}, ..., \beta_{K,t}]$ as a vector of the estimated causal effects $\beta_{s,t}$ of the exposures (features $s=1,...,K$) on the outcome (task $t$). This approach defines similar tasks as having similar exposures, but it is important to note that the assumptions of MR may not always hold in practice, for instance, due to pleiotropic variants (influencing multiple traits).
    \item Similarity using invariant causal prediction (ICP): ICP finds predictors that remain invariant across different environments. Similar to MR, we construct $\bm{e}_t = [\beta_{1,t}, ..., \beta_{K,t}]$ as a vector of the estimated causal effects. In ICP, the $\bm{z}_t$ variable is a categorical variable indicating experiment or intervention settings (environments). For example, different healthcare environments, treatment protocols, or patient subgroups based on demographic factors. ICP assumes that the conditional distribution of the outcome given the exposures does not change when intervening on variables other than the outcome via $\bm{z}_t$. Due to implementation challenges with the available software packages, we opted to use the standard ICP algorithm \cite{peters2016causal}, which makes a simplifying assumption of no hidden confounders. 
    \item A predictive baseline: We construct $\bm{e}_t = [\chi_{1,t}^2, ..., \chi_{K,t}^2]$ as a vector of chi-square test statistics, where $\chi_{s,t}^2$ represents the strength of association between feature $s$ and the outcome for task $t$. The chi-square test is used to assess the independence between each feature and the outcome, with larger values indicating stronger associations. This method does not incorporate additional information to aid with identification of causal predictors, to act as a point of comparison for the other techniques.
\end{enumerate}

\subsection{Application for Bayesian Neural Networks}
\label{ssec:bnnmethods}

For experiments we use Bayesian neural networks (BNNs) for the predictive models and a variational inference approach for estimating their posterior distributions. BNNs are suited to health data because they provide uncertainty estimates for predictions and are more robust against overfitting in scenarios with limited data \cite{polson2017deep, wilson2020case}. Using the same model class we compare a) the baseline of separate models for each task, b) the standard meta-learning approach, and c) modifications to meta-learning that account for task similarity:

\begin{enumerate}
    \item Local baselines: The variational approach assumes it is intractable to compute the posterior distribution over the neural network parameters $p(\bm{\phi}_t \mid \mathcal{D}_t)$ and instead approximates the posterior using a variational distribution $q_{\bm{\psi}_t}(\bm{\phi}_t)$ with variational parameters $\bm{\psi}_t$. Assuming a Normal distribution over the task variational parameters $N(\bm{\psi}_t| \bm{\mu}_{\bm{\psi}_t}, \bm{\sigma}^2_{\bm{\psi}_t})$, we seek variational parameters that minimize the Kullback-Leibler (KL) divergence – a measure of the difference between two probability distributions – the approximation and the true posterior:

\begin{equation}
\begin{aligned}
\underset{\bm{\psi}_t}{\operatorname{argmin}} \ \infdivkl{q_{\bm{\psi}_t}(\bm{\phi}_t)}{p(\bm{\phi}_t \mid \mathcal{D}_t)}.
\end{aligned}
\label{eq:vilocal}
\end{equation}

    This is equivalent to minimizing the negative evidence lower bound (NELBO) with respect to $\bm{\psi}_t$:

\begin{equation}
\begin{aligned}
\bm{\psi}_t^* = \underset{\bm{\psi}_t}{\operatorname{argmin}} \   \infdivkl{q_{\bm{\psi}_t}(\bm{\phi}_t)}{p(\bm{\phi}_t)} - \mathbb{E}_{q_{\bm{\psi}_t}(\bm{\phi}_t)} [ \log{p(\bm{y}_t \mid \bm{x}_t, \bm{\phi}_t)} ].
\end{aligned}
\label{eq:vilocalelbo}
\end{equation}

    We use a diagonal Normal prior distribution $p(\bm{\phi}_t) = N(\bm{\phi}_t ; \bm{0}, \bm{\sigma}^2_{\bm{\phi}_t} \bm{I})$, and employ the binary cross-entropy loss for the likelihood term, which is appropriate for the experiments with binary classification tasks in Section \ref{sec:case_study}.

    \item Meta-learning baseline: Extending (\ref{eq:vilocalelbo}) to a hierarchical model for meta-learning with global parameters $\bm{\theta}$ shared across $N$ tasks gives rise to a nested optimization objective:

\begin{equation}
\begin{aligned}
\bm{\lambda}^* = \underset{\bm{\lambda}}{\operatorname{argmin}} \ - \Big[ 
     \mathbb{E}_{q_{\bm{\lambda}}(\bm{\theta})}  \sum_{t=1}^N \big[ 
        \mathbb{E}_{q_{\bm{\psi}_t}(\bm{\phi}_t \mid \mathcal{D}_t^{(s)})} [ \log{p(\bm{y}_t^{(q)} \mid \bm{x}_t^{(q)}, \bm{\phi}_t)} ] -  \infdivkl{q_{\bm{\psi}_t}(\bm{\phi}_t \mid \mathcal{D}_t^{(s)})}{p(\bm{\phi}_t \mid \bm{\theta})} 
    \big]  \\ - \infdivkl{q_{\bm{\lambda}}(\bm{\theta})}{p(\bm{\theta})} 
\Big],
\end{aligned}
\label{eq:vimetaelbo}
\end{equation}

  where $\bm{\lambda}$ are the variational parameters for $\bm{\theta}$. This is implemented as a bi-level optimization algorithm, alternating between optimizing the meta-objective (\ref{eq:vimetaelbo}) and inner objectives of task-specific learners. An important implementation detail is the use of support $\mathcal{D}_t^{(s)}$ and query $\mathcal{D}_t^{(q)}$ splits of the task data $\mathcal{D}_t$ to aid generalizability: $\mathcal{D}_t^{(s)}$ is used to update the task-specific parameters, and the log-likelihood for updating the global parameters is computed over $\mathcal{D}_t^{(q)}$. This encourages the meta-learner to ``learn how to learn'' robust models that generalize to new data samples. The global parameters use a diagonal Normal prior distribution $p(\bm{\theta}) = N(\bm{\theta} ; \bm{0}, \bm{\sigma}^2_{\bm{\theta}} \bm{I})$ and the prior for the local parameters is meta-learned during training as $p(\bm{\phi}_t) = N(\bm{\phi}_t ; \bm{\mu}_{\bm{\theta}}, \bm{\sigma}^2_{\bm{\theta}} \bm{I})$.

    \item Meta-learning with task similarity: Finally, we extend (\ref{eq:vimetaelbo}) to estimate (\ref{eq:hbmjointposterior}) using the optimization objective
    
    \begin{equation}
    \begin{aligned}
    \bm{\lambda}^* = \underset{\bm{\lambda}}{\operatorname{argmin}} \ - \Big [ \mathbb{E}_{q_{\bm{\lambda}}(\bm{\theta})} \sum_{v=1}^{N}  \big [ \mathbb{E}_{q_{\bm{\kappa}_v}(\bm{\gamma}_v)}  \sum_{t=1}^{N} \tilde{w}_{v,t} \times \big [ \mathbb{E}_{q_{\bm{\psi}_t}(\bm{\phi}_t \mid \mathcal{D}_t^{(s)})} [  \log{p(\bm{y}_t^{(q)} \mid \bm{x}_t^{(q)}, \bm{\phi}_t)} ]  - \\  \infdivkl{q_{\bm{\psi}_t}(\bm{\phi}_t \mid \mathcal{D}_t^{(s)})}{p(\bm{\phi}_t \mid \bm{\gamma}_{v}, \bm{\theta})} \big ]  - \infdivkl{q_{\bm{\kappa}_v}(\bm{\gamma}_{v})}{p(\bm{\gamma}_{v} \mid \bm{\theta})} \big ] - \infdivkl{q_{\bm{\lambda}}(\bm{\theta})}{p(\bm{\theta})} \Big ],
    \end{aligned}
    \label{eq:vimetataskelbo}
    \end{equation}

    which introduces variational parameters $\bm{\kappa}_v$ for learning models over similar tasks. The task similarity weights $\bm{w}$ are normalized as $\tilde{w}_{v,t} = \frac{w_{v,t}}{\frac{1}{N} \sum_s w_{s,t}}$ to maintain the scale of the likelihood.

\end{enumerate}

\begin{algorithm}
\caption{Meta-learning with task similarity}\label{alg:alg1}
\begin{algorithmic}
   \STATE {\bfseries Require:} Distribution over $p(\mathcal{T})$; learning rates $\beta_1, \beta_2, \beta_3$; number of SGD updates $G_1, G_2$; temperatures $T_1, T_2, T_3$; mini-batch sizes $M_{\text{tasks}}, M_{\text{samples}}$; weights $\bm{w}$; modulating factors $\eta_1, \eta_2$; average number of similar tasks $\alpha$
   \STATE {\bfseries Initialize:} $\bm{\lambda} = (\bm{\mu_\lambda}, \bm{\sigma_\lambda^2})$ 
   \WHILE{not done}
   \STATE Sample minibatch of $M_{\text{tasks}}$ tasks $v \sim p(\mathcal{T})$ using task similarity weights $\bm{w}$ modulated by $\eta_2$
   \FOR{all $v$ in minibatch}
    \STATE {\bfseries Initialize:} $\bm{\kappa_v} = (\bm{\mu_\lambda}, \bm{\sigma_\lambda^2})$ from $\bm{\lambda}$
   \FOR{$g_1=0$ {\bfseries to} $G_1-1$}
   \STATE $S_v = \{t \in \{1,\ldots,N\} : w_{v,t} \geq c\}$, where $c$ is selected such that the average size of $S_v$ is $\alpha$ 
   \FOR{all $t$ in $S_v$}
   \STATE {\bfseries Initialize:} $\bm{\psi_t} = (\bm{\mu_\kappa}, \bm{\sigma_\kappa^2})$ from $\bm{\kappa}_v$
   \FOR{$g_2=0$ {\bfseries to} $G_2-1$}
    \STATE Sample $M_{\text{samples}}$ from $\mathcal{D}_t$ and split $\mathcal{D}_t^{(s)}, \mathcal{D}_t^{(q)}$  
   \STATE $\bm{\psi}_t^{(g_2+1)} \leftarrow \bm{\psi}_t^{(g_2)} - \beta_1 \nabla_{\bm{\psi}_t^{(g_2)}} \mathcal{L}_1 (\bm{\psi}_t^{(g_2)}, \mathcal{D}_t^{(s)})$
   \ENDFOR
   \ENDFOR
   \STATE $\bm{\kappa}_v^{(g_1+1)} \leftarrow \bm{\kappa}_v^{(g_1)} - \beta_2 \nabla_{\bm{\kappa}_v^{(g_1)}}  \mathcal{L}_2 (\bm{\kappa}_v^{(g_1)}, \mathcal{D}^{(q)}, \bm{w}, \eta_1)$
   \ENDFOR
   \ENDFOR
   \STATE $\bm{\lambda} \leftarrow \bm{\lambda} - \beta_3 \nabla_{\bm{\lambda}} \big [ \frac{1}{M_{\text{tasks}}} \sum_t \mathcal{L}_2 (\bm{\kappa}_t^{(G_1)}, \mathcal{D}^{(q)}, \bm{w}, \eta_1) - \frac{1}{N_{\text{tasks}}} \log {p(\bm{\theta})} + \log{q_{\bm{\lambda}}(\bm{\theta})} \times \frac{T_3}{N_{\text{tasks}}} \big ]$
   \ENDWHILE
\end{algorithmic}
\end{algorithm}

Algorithm \ref{alg:alg1} shows our full algorithm for meta-learning with task similarity (the baseline methods are simpler variations of this), which extends previous work on amortized variational inference for Bayesian meta-learning \cite{ravi2018amortized}. Meta-training is carried out according to Algorithm \ref{alg:alg1} to derive the shared model $\bm{\lambda}$ across training tasks, which is used to derive task-specific models (for training or new tasks) $\bm{\psi}_t$, also according to Algorithm \ref{alg:alg1}.

Our code implementations of these methods use Python and functional PyTorch via the Posteriors Python library \cite{duffield2024scalable}, which we have modified to handle the higher-order gradients required for meta-learning. We apply standard strategies for improving the stability of the optimization procedure and reducing computational complexity. Specifically, the prior and likelihood terms in the optimization objectives have been scaled to account for mini-batching and tempered posteriors are used by introducing temperature hyperparameters $T_1, T_2, T_3$ that control how much emphasis is placed on the prior distributions relative to the likelihood terms:

\begin{equation}
\begin{aligned}
\mathcal{L}_1 (\bm{\psi}_t, \mathcal{D}_t^{(s)}) = - \big [ \frac{1}{M_{\text{samples}}} \sum_i \log{p(\bm{y}_{t,i}^{(s)} \mid \bm{x}_{t,i}^{(s)}, \bm{\phi}_t)} + \\ \frac{1}{N_{\text{samples}}} \log {p(\bm{\phi}_t \mid \bm{\gamma}_v, \bm{\theta})} -  \log{q_{\bm{\psi}_t}(\bm{\phi}_t \mid \mathcal{D}_t^{(s)})} \times \frac{T_1}{N_{\text{samples}}} \big ],
\end{aligned}
\label{eq:viloss1}
\end{equation}
\begin{equation}
\begin{aligned}
\mathcal{L}_2 (\bm{\kappa}_v, \mathcal{D}^{(q)}, \bm{w}, \eta_1) = -  \big [  \frac{1}{N_{\text{tasks},v}} \sum_t \tilde{w}_{v,t}^{\eta_1} \times \mathcal{L}_1 (\bm{\psi}_t, \mathcal{D}_t^{(q)}) + \\ \frac{1}{N_{\text{tasks},v}} \log {p(\bm{\gamma}_v \mid \bm{\theta})} - \log{q_{\bm{\kappa}_v}(\bm{\gamma}_v)} \times \frac{T_2}{N_{\text{tasks},v}} \big ],
\end{aligned}
\label{eq:viloss2}
\end{equation}
where $M_{\text{samples}}$ is the number of samples in a mini-batch, $N_{\text{samples}}$ is the total number of samples, $M_{\text{tasks}}$ is the number of tasks in a mini-batch, $N_{\text{tasks},v}$ is the number of activated tasks for task $v$, and $N_{\text{tasks}}$ is the total number of tasks. A set of activated tasks $S_v$ is defined for each task $v$, based on a similarity threshold $c$ that is determined by a pre-specified average number of activated tasks $\alpha$. This constrains the computational complexity of task similarity-based meta-learning for a large number of tasks. Notably, when $G_1 = 1$, $G_2 = G$, and $S_v = \{ v \}$, the algorithm reduces to standard meta-learning with $G$ inner updates, providing a clear connection to simpler approaches.

In addition to the average number of activated tasks $\alpha$, modulating factors $\eta_1$ and $\eta_2$ are introduced to tune how the model utilizes similar tasks; $\eta_1$ controls how task similarity is used in the optimization objective (\ref{eq:viloss2}) and $\eta_2$ controls how task similarity is used in mini-batching (i.e., whether more similar tasks are observed more frequently during training). Modulating factors are applied to task-similarity weights as $\bm{w}^{\eta_i}$, where higher modulation gives higher weighting to the most similar tasks. The values of $\alpha, \eta_1, \eta_2$ and all other hyperparameters are determined using Bayesian optimization \cite{snoek2012practical}, allowing the algorithm to automatically discover how to best utilize task similarity.

\section{Experiments and results}
\label{sec:case_study}

\subsection{Datasets}
\label{ssec:datasets}

For experiments we utilize the UK Biobank and FinnGen datasets for a stroke prediction case study (summarized in Table \ref{tab:datasummary}). We chose stroke because (1) it has inter- and intra-variability due to distinct subtypes (e.g., ischemic stroke, hemorrhagic stroke) and patient-level differences; (2) can benefit from knowledge transfer across related tasks because there is limited data for specific subtypes, but they share risk factors with other cardiovascular and metabolic diseases. 

The UK Biobank dataset is regarded as the training population, and the evaluations are carried out for new patients from both the UK Biobank dataset and the separate FinnGen dataset. The UK Biobank dataset \cite{bycroft2018uk} consists of 223,756 individuals with comprehensive primary care and hospital records to create a longitudinal dataset. The feature set comprises age, sex, and sequences of 534 diagnosis and medication codes (aggregated at yearly intervals, prior to 2011). The machine learning tasks are binary classification tasks for predicting future outcomes (following 2011) for different types of stroke and related conditions (ICD-10 codes). The meta-learning algorithm is trained for 25 tasks in total, including six target tasks for different types of stroke: G45 (TIA, transient cerebral ischemic attack), I60 (SAH, subarachnoid hemorrhage), I61 (ICH, intracerebral hemorrhage), I62 (SDH/EDH, subdural/epidural hemorrhage), I63 (IS, ischemic stroke), and I64 (CVA, cerebrovascular accident, or stroke, undefined). A range of related conditions are included as the remaining tasks, including E11 (T2D, type 2 diabetes), I10 (HTN, essential hypertension), and I12 (CKD, hypertensive chronic kidney disease). The FinnGen dataset \cite{kurki2023finngen} consists of 472,845 individuals and is prepared similarly to UK Biobank, matching equivalent variables where possible, resulting in 25 tasks including 1 target task for stroke.

For causal task similarity analyses, we utilize additional metadata including summary statistics from genome-wide association studies (GWAS) and fine-mapping analyses for the UK Biobank and FinnGen cohorts \cite{nealelab, kurki2023finngen}. While we do not incorporate individual-level genetics data directly as features for machine learning, as such data is often unavailable in clinical EHR systems, we aim to utilize publicly available genetics statistics for identifying tasks with similar disease mechanisms.

\begin{table}[ht]
\centering
\caption{Summary Statistics for Data Used in Experiments}
\label{tab:datasummary}
\setlength{\tabcolsep}{3pt}
\begin{tabular}{p{160pt}p{140pt}p{140pt}}
\hline
 & \textbf{UK Biobank (UKB)} & \textbf{FinnGen} \\
\hline
Total Participants & 223,756 & 472,845 \\
\hline
$<$20 years & 0 (0.0\%) & 13,323 (2.8\%) \\
20-29 years & 0 (0.0\%) & 40,992 (8.7\%) \\
30-39 years & 0 (0.0\%) & 58,797 (12.4\%) \\
40-49 years & 34,559 (15.4\%) & 66,098 (13.9\%) \\
50-59 years & 67,819 (30.3\%) & 96,405 (20.4\%) \\
60-69 years & 99,151 (44.3\%) & 105,206 (22.2\%) \\
70+ years & 22,227 (9.9\%) & 91,248 (19.3\%) \\
\hline
Male & 100,826 (45.1\%) & 204,361 (43.2\%) \\
Female & 122,930 (54.9\%) & 268,484 (56.8\%) \\
\hline
\multicolumn{3}{c}{\textbf{Target Task Statistics}} \\
\hline
 & Male & Female \\
\hline
UKB G45 (TIA) & 1337 (1.33\%) & 1109 (0.90\%) \\
UKB I60 (SAH) & 130 (0.13\%) & 266 (0.22\%) \\
UKB I61 (ICH) & 361 (0.36\%) & 295 (0.24\%) \\
UKB I62 (SDHs/EDHs) & 251 (0.25\%) & 141 (0.11\%) \\
UKB I63 (IS) & 1764 (1.75\%) & 1192 (0.97\%) \\
UKB I64 (CVA) & 672 (0.67\%) & 450 (0.37\%) \\
FinnGen C\_STROKE & 28,482 (13.94\%) & 21,509 (8.01\%) \\
\hline
 & 40-59 years & 60+ years \\
\hline
UKB G45 (TIA) & 531 (0.52\%) & 1915 (1.58\%) \\
UKB I60 (SAH) & 123 (0.12\%) & 273 (0.22\%) \\
UKB I61 (ICH) & 138 (0.13\%) & 518 (0.43\%) \\
UKB I62 (SDHs/EDHs) & 65 (0.06\%) & 327 (0.27\%) \\
UKB I63 (IS) & 588 (0.57\%) & 2368 (1.95\%) \\
UKB I64 (CVA) & 182 (0.18\%) & 940 (0.77\%) \\
FinnGen C\_STROKE & 13,275 (20.09\%) & 34,702 (17.66\%) \\
\hline
\end{tabular}
\end{table}

\subsection{Experiment setup}

The experiments aim to answer the following questions: a) Does meta-learning generalize better than separate models for each task, does negative transfer occur, and what factors contribute to this? b) How similar are the tasks and do different methods for identifying similar tasks produce different results? c) Can similar tasks be leveraged to mitigate negative transfer and improve the generalizability of meta-learning to new patients (from the training and new populations)? 

The experiments use a BNN model based on the Long Short-Term Memory (LSTM) architecture, which has been widely applied to handle the temporal aspects of EHR data \cite{morid2023time}. Specifically, the architecture consists of an LSTM layer for the sequence data (diagnosis and medication variables), and the outputs of the final time step are concatenated with a fully connected layer's output for the remaining features, before being passed through a final output layer for binary classification. 

The meta-learning models and baselines are trained using the UK Biobank dataset. Specifically, non-test patient samples are split into $k$ folds of $M_{\text{train}}$ train samples and $M_{\text{val}}$ validation samples for model training and hyperparameter tuning using cross-validation and Bayesian optimization. Table \ref{tab:hyperparameters} (Appendix) details the hyperparameter search space and other settings used in experiments. Non-test patient samples are also used for adapting models to specific tasks, for both the UK Biobank and FinnGen datasets. The evaluation criteria assess the methods' generalizability to a further $M_{\text{test}}$ patients and report binary classification metrics: Area Under the Receiver Operating Characteristic curve (AUROC), Area Under the Precision-Recall Curve (AUPRC). Post-training, we conducted additional analyses to gain deeper insights into model generalizability behavior, comparing: (1) feature importance estimates using the integrated gradients method \cite{sundararajan2017axiomatic}; (2) patient characteristics associated with model confidence using logistic regression analysis. The FinnGen experiments used an AMD EPYC 7B12 processor with 16 CPUs and 128 GB RAM, and the UK Biobank experiments used a supercomputing cluster with 56 Tesla A100 80GB RAM NVIDIA GPUs, which allowed for more extensive analyses.

\subsection{Results of baseline meta-learning analysis}
\label{ssec:resultsbaselines}

We first investigate generalizability and negative transfer in the baseline methods for the training dataset (UK Biobank). The results summarized in Table \ref{tab:predictiveresults} show that the meta-learning baseline outperforms the local baselines (separate models for each task) for 4 out of 6 UK Biobank target tasks: the largest improvement was for task I63/IS (4.32\% improvement in AUROC), followed by I64/CVA (2.27\%), I61/ICH (1.91\%) and G45/TIA (1.04\%), while meta-learning performed worse for task I60/SAH (6.66\% decrease) and I62/SDH/EDH (1.81\%). This indicates negative transfer for 2 target tasks: subarachnoid hemorrhage (I60, SAH), and to a lesser extent, subdural/epidural hemorrhage (I62, SDH/EDH). 

\begin{figure*}[!t]
\centerline{\includegraphics[width=0.9\linewidth]{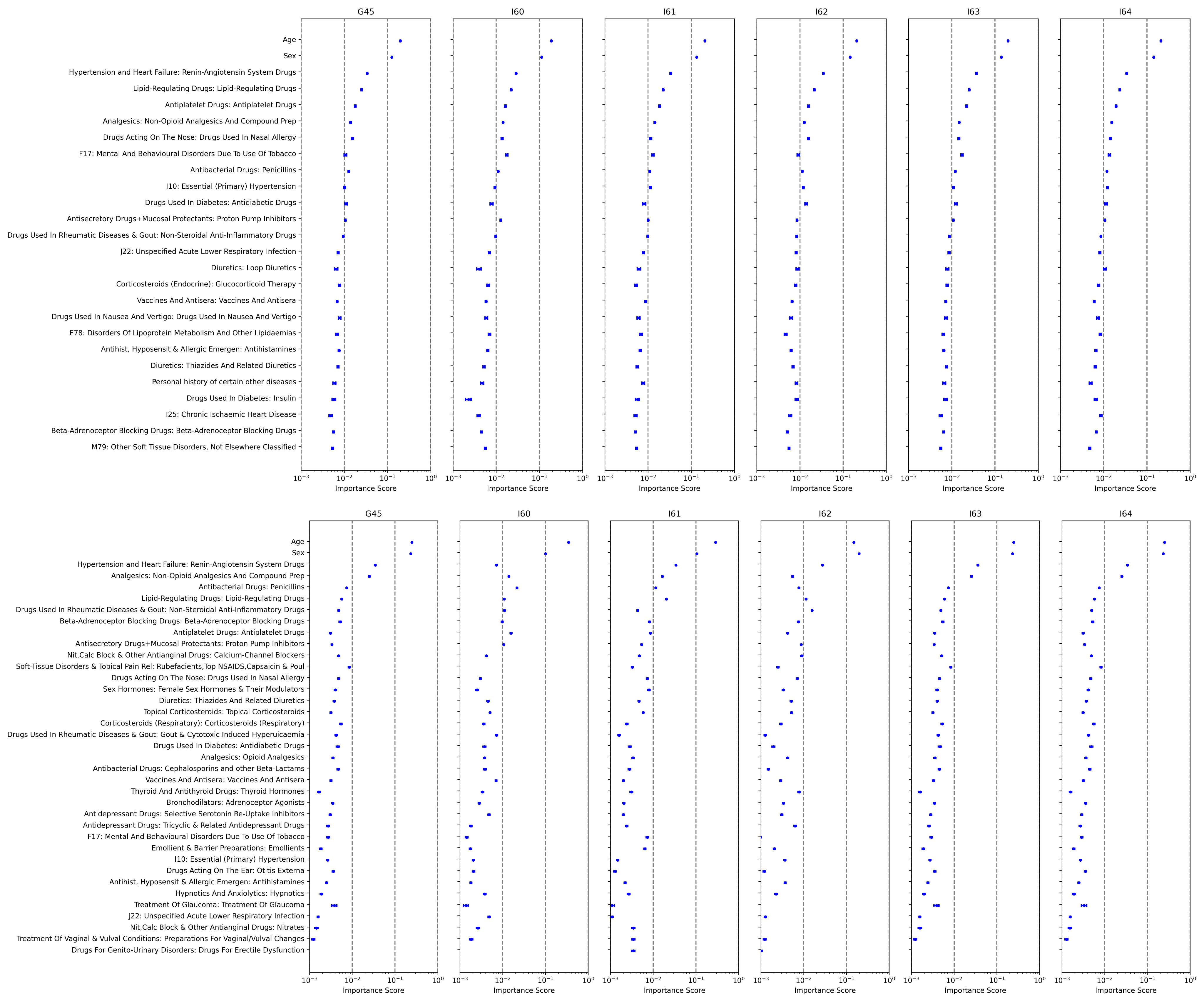}}
\caption{Feature importance scores for different stroke types (G45, I60, I61, I62, I63, I64) calculated using the integrated gradients approach. Top row: Meta-learning baseline model. Bottom row: Meta-learning with task similarity, using the Mendelian Randomization (MR) approach. Features shown on the y-axis were ranked in the top 20 for at least one stroke type. The x-axis represents the importance score on a logarithmic scale.}
\label{fig:feat_imp}
\end{figure*}

We next analyze if shortcut learning of (non-causal) spurious associations contributes to this negative transfer. Figure \ref{fig:feat_imp} shows the results from a feature importance analysis, which can reveal over-reliance on specific data features (including potentially spurious associations). The meta-learning models rely on similar features across all target tasks, with the same top-ranking features: age, sex, and the medication features renin-angiotensin system (RAS) drugs and lipid-regulating drugs. These drugs are related to risk factors for stroke: RAS drugs are primarily used to manage hypertension, a significant risk factor for all strokes \cite{Andone2022NeuroprotectionIS}; lipid-regulating drugs (particularly statins) are used to lower cholesterol, but the relationship between lipids and stroke varies by stroke type \cite{Yaghi2015LipidsAC}. While other high-ranking features are related to known risk factors of stroke (hypertension, tobacco use, disorders of lipoprotein metabolism), drugs related to primary or secondary prevention (diuretics, beta-blockers, diabetes drugs, antiplatelet drugs) or possible increased risk of stroke (corticosteroids, analgesics), others have complex or inconclusive relationships with stroke (allergy drugs, penicillins, proton pump inhibitors) \cite{Tuttolomondo2014StrokeSA, Schulz2013DrugTI}. Table \ref{tab:feature_importance_correlation} quantitatively compares feature importance across pairs of task models using Spearman rank correlation. The values ranged from 0.89 for I62 (SDH/EDH) and I64 (CVA), to 0.96 for G45 (TIA) and I63 (IS). The high correlation values indicate that even the least similar models had substantial overlap in important features. This tendency of the meta-learned models to learn common patterns across all tasks even where they may not be relevant for all stroke subtypes may contribute to the observed negative transfer.

\begin{figure*}[!t]
\centerline{\includegraphics[width=0.99\linewidth]{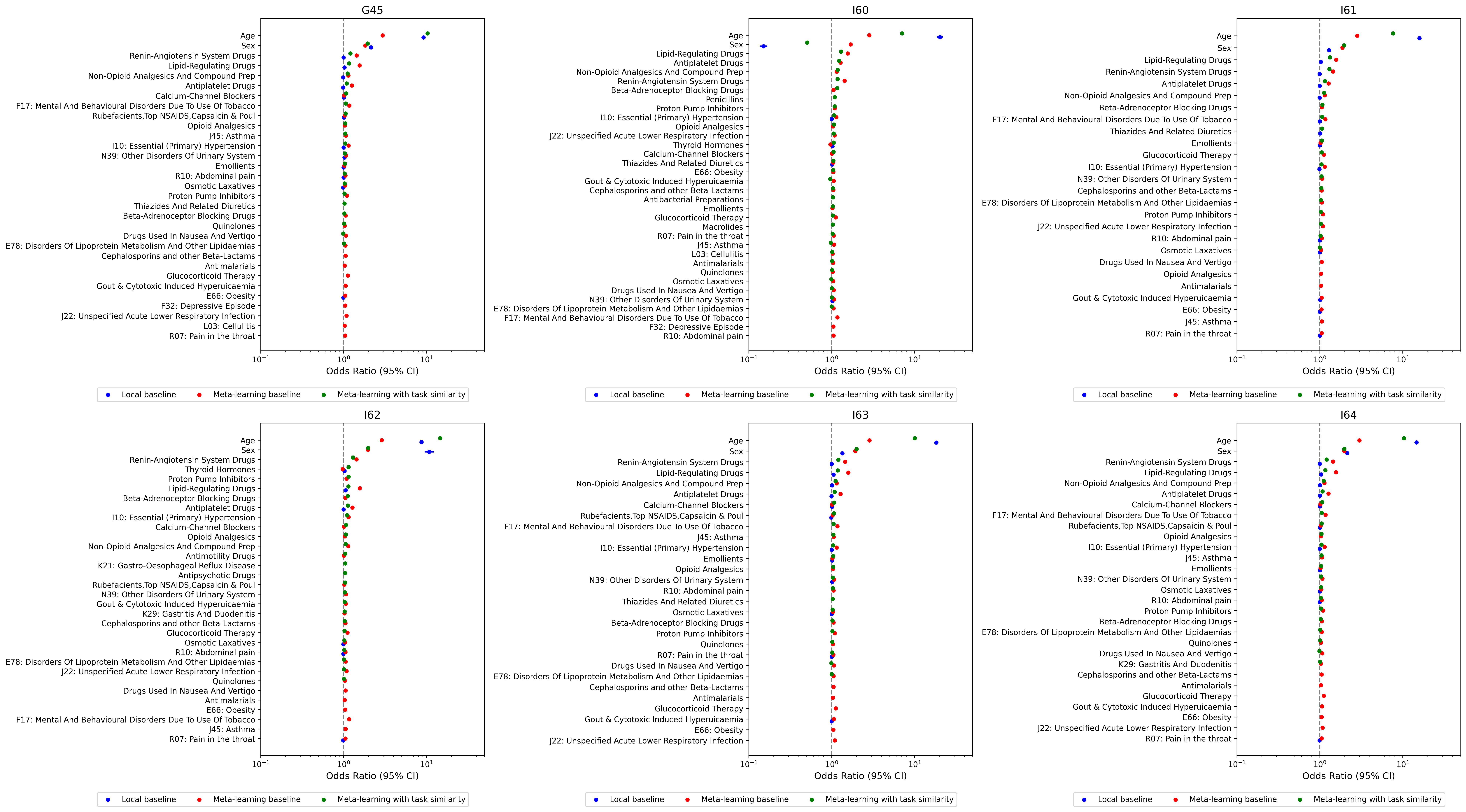}}
\caption{Comparison of odds ratios for local baseline (blue), meta-learning baseline (red), and meta-learning with MR-based task similarity (green) for predicting different stroke types (G45, I60, I61, I62, I63, I64) with high confidence (above median entropy). The plots show odds ratios with 95\% confidence intervals for various factors, which were selected based on their significance (p $<$ 0.001) and positive association (odds ratio $>$ 1) in at least one model.}
\label{fig:lr_odds_ratios}
\end{figure*}

We examine what factors influence generalizability performance of the models. Figure \ref{fig:lr_odds_ratios} reports a logistic regression analysis of patient characteristics associated with high-confidence predictions of stroke (based on uncertainty estimates from the models). 
The factors associated with high-confidence prediction are similar across tasks: older age (average OR 2.894, 95\% CI: 2.800, 2.991), male sex (average OR 1.880, 95\% CI: 1.824, 1.937), and medications including renin-angiotensin system drugs (average OR 1.612, 95\% CI: 1.564, 1.662). This behavior can be beneficial when stroke risk is explained by similar factors, but can contribute to poor generalizability for patient groups with less common risk factors. This is observed for task I60 (SAH), which has different patient demographics to the other tasks analyzed (Table \ref{tab:datasummary}): I60 (SAH) is more common in females (0.22\%) than in males (0.13\%). The local model associates female sex with high-confidence predictions of stroke, whereas the meta-learning model behaves similar to the meta-learning models of other tasks and makes more confident predictions for male patients (Figure \ref{fig:lr_odds_ratios}).

\begin{landscape}
\begin{table*}[t]
\centering
\caption{Comparison of Machine Learning Models for UK Biobank and Adaptation to FinnGen Data}
\label{tab:predictiveresults}
\scriptsize
\setlength{\tabcolsep}{2pt}
\begin{tabular}{@{}p{2.5cm}|cccccc|c@{}}
\hline
& \multicolumn{6}{c|}{\textbf{UK Biobank}} & \textbf{FinnGen} \\
\cline{2-8}
\textbf{Method} & G45 (TIA) & I60 (SAH) & I61 (ICH) & I62 (SDHs/EDHs) & I63 (IS) & I64 (CVA) & C\_STROKE \\
\hline
& \multicolumn{7}{c}{\textbf{AUROC}} \\
\hline
Local Baselines & 0.672 [0.669-0.675] & 0.601 [0.591-0.612] & 0.682 [0.677-0.689] & 0.720 [0.715-0.727] & 0.694 [0.689-0.698] & 0.704 [0.699-0.709] & - \\
Meta-learning Baseline & 0.679 [0.675-0.685] & 0.561 [0.554-0.570] & 0.695 [0.685-0.704] & 0.707 [0.699-0.714] & 0.724 [0.721-0.728] & 0.720 [0.716-0.724] & 0.738 [0.731-0.743] \\
Meta-learning with Task Similarity: & & & & & & & \\
\quad DAG Similarity & \textbf{0.688} [0.685-0.693] & 0.581 [0.573-0.589] & \textbf{0.702} [0.695-0.710] & \textbf{0.727} [0.720-0.735] & 0.723 [0.720-0.727] & \textbf{0.723} [0.719-0.726] & \textbf{0.754} [0.743-0.761] \\
\quad MR Similarity & \textbf{0.688} [0.685-0.690] & \textbf{0.616} [0.607-0.628] & \textbf{0.705} [0.699-0.715] & \textbf{0.730} [0.723-0.738] & 0.724 [0.719-0.729] & \textbf{0.726} [0.719-0.732] & \underline{\textbf{0.764}} [0.754-0.775] \\
\quad ICP Similarity & \textbf{0.687} [0.684-0.692] & 0.576 [0.567-0.584] & \underline{\textbf{0.716}} [0.704-0.726] & \underline{\textbf{0.736}} [0.729-0.750] & \textbf{0.728} [0.725-0.730] & \underline{\textbf{0.730}} [0.725-0.735] & \underline{\textbf{0.767}} [0.758-0.774] \\
\quad Chi-2 Similarity & \underline{\textbf{0.689}} [0.685-0.691] & 0.555 [0.548-0.560] & \textbf{0.704} [0.700-0.713] & \textbf{0.732} [0.724-0.739] & 0.722 [0.721-0.724] & \underline{\textbf{0.731}} [0.727-0.737] & \underline{\textbf{0.750}} [0.746-0.756] \\
\hline
& \multicolumn{7}{c}{\textbf{AUPRC}} \\
\hline
Local Baselines & 0.020 [0.020-0.021] & 0.002 [0.002-0.002] & 0.005 [0.005-0.005] & 0.005 [0.005-0.006] & 0.025 [0.025-0.026] & 0.011 [0.010-0.012] & - \\
Meta-learning Baseline & 0.021 [0.021-0.022] & 0.002 [0.002-0.002] & 0.007 [0.006-0.007] & 0.004 [0.004-0.005] & 0.031 [0.031-0.032] & 0.012 [0.012-0.013] & 0.124 [0.120-0.128] \\
Meta-learning with Task Similarity: & & & & & & & \\
\quad DAG Similarity & \textbf{0.022} [0.022-0.023] & 0.002 [0.002-0.002] & \underline{\textbf{0.011}} [0.008-0.013] & 0.004 [0.004-0.005] & \underline{\textbf{0.032}} [0.032-0.033] & \textbf{0.013} [0.012-0.013] & \underline{\textbf{0.139}} [0.132-0.144] \\
\quad MR Similarity & \textbf{0.022} [0.021-0.022] & 0.002 [0.002-0.002] & \underline{\textbf{0.009}} [0.008-0.010] & 0.005 [0.004-0.005] & \textbf{0.032} [0.031-0.033] & 0.012 [0.011-0.013] & \textbf{0.129} [0.125-0.133]  \\
\quad ICP Similarity & \underline{\textbf{0.023}} [0.022-0.023] & 0.002 [0.002-0.002] & \underline{\textbf{0.013}} [0.012-0.014] & \textbf{0.006} [0.005-0.006] & \textbf{0.032} [0.031-0.032] & \textbf{0.013} [0.012-0.013] & \underline{\textbf{0.138}} [0.134-0.143] \\
\quad Chi-2 Similarity & \textbf{0.022} [0.022-0.023] & 0.002 [0.002-0.002] & \underline{\textbf{0.009}} [0.008-0.009] & \textbf{0.006} [0.005-0.006] & 0.031 [0.030-0.032] & \textbf{0.013} [0.012-0.013] & 0.121 [0.117-0.127]  \\
\hline
\multicolumn{8}{l}{\footnotesize AUROC: Area Under the Receiver Operating Characteristic curve and 95\% confidence intervals} \\
\multicolumn{8}{l}{\footnotesize AUPRC: Area Under the Precision-Recall Curve and 95\% confidence intervals} \\
\multicolumn{8}{l}{\footnotesize Bold: Improvement over both baselines. Underline: Significant improvement, based on confidence intervals.} \\
\end{tabular}
\end{table*}
\end{landscape}

\subsection{Task similarity analysis}
\label{ssec:tasksim}

\begin{figure*}[!t]
\centerline{\includegraphics[width=0.9\linewidth]{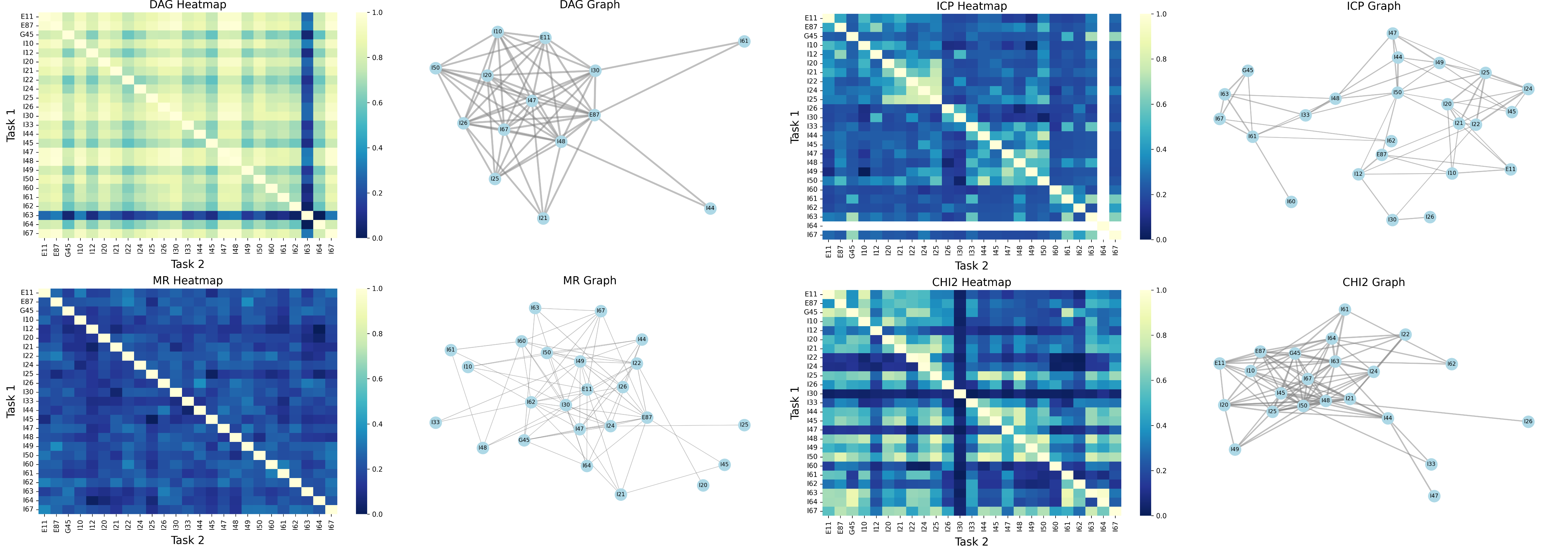}}
\caption{Comparison of task similarity analyses using four different measures: Directed Acyclic Graph (DAG), Mendelian Randomization (MR), Invariant Causal Prediction (ICP), and Chi-square test (CHI2). Heatmaps on the left show pairwise task similarities (lighter colors indicate higher similarity) and graphs on the right visualize the relationships between tasks (edge inclusion defined as top 20\% most similar tasks).}
\label{fig:tasksimilarity}
\end{figure*}

Figure \ref{fig:tasksimilarity} reports task similarity analyses for the UK Biobank data using the measures described in Section \ref{ssec:tasksimmethods}. Details for implementing these analyses are provided in Appendix \ref{apx:expsettings}, which focused on diagnosis features (excluding medications due to the lack of appropriate auxiliary data needed for the causal inference techniques). The different measures produced varying results, highlighting that task similarity can differ significantly depending on the underlying assumptions of the causal model being considered. The MR and ICP measures produced more similar results to each other (Frobenius distance 4.3 between task similarity matrices) than the CHI2 baseline (Frobenius distance of 6.7 and 5.5 for MR and ICP, respectively), and were the most successful at differentiating tasks based on underlying biological mechanisms. Table \ref{tab:stroke-similarity} shows the top-ranked most similar tasks for each stroke task. The ICP measure found high similarity between ischemic strokes G45 (TIA) and I63 (IS), and between hemorrhagic strokes I60 (SAH), I61 (ICH) and I62  (SDH/EDH); however, the ICP measure also identified high similarity between I61 (ICH) and I63 (IS), and produced inconclusive results for task I64 (CVA), which is a catch-all code for strokes that are not clearly defined and therefore may not have an invariant set of causal predictors. The MR method also identified relationships distinguishing ischemic and hemorrhagic strokes: G45 (TIA) most similar to I63 (IS) and I64 (CVA), and I61 (ICH) most similar to I60 (SAH). The DAG measure identified more similarity with non-stroke diagnoses than other stroke types, while the CHI2 measure did not show a clear separation of known stroke types. These measures may have inferred other forms of task similarity from the data. The DAG measure stood out as the most distinct, with Frobenius distances of 12, 12, and 14 compared to CHI2, ICP, and MR, respectively. This is likely because DAG is the only multivariate approach, while the other measures are based on bivariate (exposure-outcome) relationships.

\begin{table}[ht]
\centering
\caption{Top-4 Most Similar Tasks for Different Stroke Types Using Various Methods}
\label{tab:stroke-similarity}
\resizebox{0.8\textwidth}{!}{%
\begin{tabular}{|c|c|c|c|c|c|}
\hline
\multirow{2}{*}{\textbf{Stroke}} & \multirow{2}{*}{\textbf{Rank}} & \textbf{CHI2} & \textbf{DAG} & \textbf{ICP} & \textbf{MR} \\
\textbf{Type} & & \textbf{Similarity} & \textbf{Similarity} & \textbf{Similarity} & \textbf{Similarity} \\
\hline
\multirow{4}{*}{G45 (TIA)} 
 & 1 & I63 (0.87) & E87 (0.81) & I63 (0.70) & I63 (0.31) \\
 & 2 & I64 (0.86) & I20 (0.80) & I67 (0.66) & I64 (0.28) \\
 & 3 & I50 (0.69) & I47 (0.80) & I61 (0.49) & I62 (0.27) \\
 & 4 & I48 (0.68) & I67 (0.79) & I24 (0.30) & I30 (0.26) \\
\hline
\multirow{4}{*}{I60 (SAH)} 
 & 1 & I45 (0.50) & I30 (0.86) & I61 (0.55) & I61 (0.33) \\
 & 2 & I50 (0.46) & E87 (0.85) & I62 (0.33) & I22 (0.31) \\
 & 3 & I26 (0.44) & I48 (0.85) & I67 (0.29) & I67 (0.29) \\
 & 4 & I67 (0.42) & I47 (0.84) & I21 (0.25) & I49 (0.28) \\
\hline
\multirow{4}{*}{I61 (ICH)} 
 & 1 & I63 (0.88) & I30 (0.88) & I63 (0.68) & I60 (0.33) \\
 & 2 & I64 (0.84) & E87 (0.87) & I67 (0.61) & I48 (0.26) \\
 & 3 & G45 (0.75) & I67 (0.87) & I60 (0.55) & I64 (0.24) \\
 & 4 & I62 (0.59) & I47 (0.86) & I33 (0.52) & I62 (0.24) \\
\hline
\multirow{4}{*}{I62 (SDHs/EDHs)} 
 & 1 & I64 (0.61) & I30 (0.82) & I67 (0.43) & I67 (0.30) \\
 & 2 & I63 (0.59) & E87 (0.82) & I61 (0.36) & I47 (0.29) \\
 & 3 & G45 (0.46) & I48 (0.81) & I60 (0.33) & E11 (0.28) \\
 & 4 & I33 (0.34) & I47 (0.81) & I26 (0.32) & I10 (0.27) \\
\hline
\multirow{4}{*}{I63 (IS)} 
 & 1 & I50 (0.80) & I48 (0.33) & G45 (0.70) & I26 (0.32) \\
 & 2 & I48 (0.79) & I10 (0.33) & I61 (0.68) & G45 (0.29) \\
 & 3 & I67 (0.77) & E87 (0.32) & I67 (0.64) & I67 (0.28) \\
 & 4 & I44 (0.67) & I47 (0.32) & I33 (0.57) & I30 (0.25) \\
\hline
\multirow{4}{*}{I64 (CVA)} 
 & 1 & I63 (0.95) & E87 (0.80) & None & I26 (0.28) \\
 & 2 & I50 (0.69) & I48 (0.79) & None & G45 (0.27) \\
 & 3 & I67 (0.60) & I47 (0.79) & None & I50 (0.26) \\
 & 4 & I45 (0.59) & I30 (0.79) & None & I30 (0.26) \\
\hline
\multicolumn{6}{l}{\small Ranking of task similarity and similarity score for corresponding method in brackets}  \\
\multicolumn{6}{p{14cm}}{\small ICD-10 Code Descriptions: G45: Transient cerebral ischemic attacks and related syndromes, 
I60: Subarachnoid hemorrhage, I61: Intracerebral hemorrhage, I62: Other nontraumatic intracranial hemorrhage, 
I63: Cerebral infarction, I64: Stroke, not specified as hemorrhage or infarction, I10: Essential (primary) hypertension, 
I20: Angina pectoris, I21: Acute myocardial infarction, I22: Subsequent myocardial infarction, 
I24: Other acute ischemic heart diseases, I26: Pulmonary embolism, I30: Acute pericarditis, 
I33: Acute and subacute endocarditis, I44: Atrioventricular and left bundle-branch block, 
I45: Other conduction disorders, I47: Paroxysmal tachycardia, I48: Atrial fibrillation and flutter, 
I49: Other cardiac arrhythmias, I50: Heart failure, I67: Other cerebrovascular diseases, 
E11: Type 2 diabetes mellitus, E87: Other disorders of fluid, electrolyte and acid-base balance}
\end{tabular}
}
\end{table}

\subsection{Meta-learning method comparison and generalizability results}

Table \ref{tab:predictiveresults} compares task similarity in Bayesian meta-learning for four measures (DAG, MR, ICP, CHI2) with the local and meta-learning baselines. Compared to local and meta-learning baselines, task similarity-based methods produced comparable or better generalizability performance (AUROC and AUPRC) across all tasks. The task similarity-based approach was also effective for mitigating the negative transfer observed in the baseline results for tasks I60 (SAH) and I62  (SDH/EDH): the ICP approach achieved the best result for task I62 (SDH/EDH) (AUROC of 0.736, 95\% CI: 0.729-0.750), and the MR approach showed an improvement for task I60 (SAH) (AUROC of 0.616, 95\% CI: 0.607-0.628). 

In addition to the UK Biobank experiments, which assess generalizability to unseen patients from the UK Biobank population, we also assessed generalizability to patients from a separate dataset (FinnGen). Similar to the UK Biobank experiments, the meta-learning models were trained using the UK Biobank dataset, but the adaptation part for evaluation was done using tasks from the FinnGen dataset. The ICP approach had the best result for this (AUROC of 0.767, 95\% CI: 0.758-0.774, compared to the meta-learning baseline AUROC of 0.738, 95\% CI: 0.731-0.743), indicating that a task similarity-based approach also aids generalizability to new patient populations (Table \ref{tab:predictiveresults}). 

Next, we examine how different types of task similarity affect these results. No task similarity measure consistently outperformed others across all tasks and datasets examined, indicating that different measures may be more beneficial depending on the task. The best performing measures were ICP for the FinnGen stroke task and UK Biobank tasks I61 (ICH), I62  (SDH/EDH) and I63 (IS); CHI2 for tasks G45 (TIA) and I64 (CVA), and MR for task I60 (SAH). Table \ref{tab:bayesopresults} summarizes from the Bayesian optimization results how task similarity was utilized by each model. All task similarity measures favored learning across multiple tasks, and the optimal number of similar tasks and modulating factors varied by method. Averaged across the highest-scoring trials, the DAG similarity approach utilized an average of 5.75 (SD = 0.50) similar tasks, MR used 4.00 (SD = 1.83), ICP used 5.00 (SD = 0.82), and CHI2 used 5.00 (SD = 1.15). 

Finally, we consider if task similarity alleviates the shortcut learning observed in the baseline experiments. Figures \ref{fig:lr_odds_ratios} and \ref{fig:feat_imp} provide an illustrative analysis of the feature importance estimates and patient characteristics for the MR model. Compared to the meta-learning baseline, the MR approach has more variation in important features across tasks, suggesting that task similarity-based meta-learning better reflects the true intra-task variability (Figure \ref{fig:feat_imp} and rank correlation results in Table \ref{tab:feature_importance_correlation}). While there were common important features across all stroke types which also overlap with the meta-learning baseline model, there was more variation in important features among hemorrhagic stroke tasks (I60/SAH, I61/ICH and I62/SDH/EDH). The patient characteristics associated with confident predictions of stroke were similar compared to the meta-learning baseline (Figure \ref{fig:lr_odds_ratios}), but a significant difference was observed for task I60 (SAH): the meta-learning baseline was more likely to predict stroke in male patients (OR 1.695, 95\% CI: 1.648, 1.743), whereas the task similarity-based meta-learning method was more likely to predict stroke in female patients (OR 0.508, 95\% CI: 0.496, 0.520), which is more consistent with the patient demographics of this task (Table \ref{tab:datasummary}). These results indicate that task similarity-based meta-learning methods can give rise to more favorable generalizability behaviors.

\section{Discussion and conclusion}
\label{sec:conclusion}

This work investigated the generalizability challenges in meta-learning for supervised learning tasks from large-scale EHR data. The main methodological contribution was a Bayesian meta-learning approach that models similarity between causal mechanisms of the tasks. Experiments for stroke prediction tasks with the UK Biobank and FinnGen datasets showed that this reduces negative transfer in meta-learning and improves generalizability of predictions to new patients, compared to local (single-task) and meta-learning baselines. Specifically, the generalizability gains were improvements in AUROC and AUPRC for new patients - from both the training (UK Biobank) patient population and a separate (FinnGen) population - and negative transfer (worse than single-task performance in standard meta-learning) was alleviated for the two UK Biobank stroke tasks (subarachnoid hemorrhage, subdural/epidural hemorrhage) in which it was observed (Table \ref{tab:predictiveresults}).

A comparison of four measures (CHI2, DAG, ICP, MR) for quantifying similar tasks showed that some measures performed better on specific tasks (i.e., ICP for most UK Biobank tasks and the FinnGen task), but there was no single measure that outperformed others across all tasks (Table \ref{tab:predictiveresults}). Our analysis used hyperparameter tuning techniques (via Bayesian optimization) to automatically determine how task relatedness is best utilized (Table \ref{tab:bayesopresults}), which proved an effective approach for identifying models that performed comparably or better than the baselines. Further task similarity analyses indicated that different measures captured different aspects of task similarity (Figure \ref{fig:tasksimilarity} and Table \ref{tab:stroke-similarity}). Specifically, the MR and ICP measures differentiated tasks with shared underlying biological mechanisms (ischemic and hemorrhagic strokes), and this form of task similarity was shown to be advantageous for improving generalizability (Table \ref{tab:predictiveresults}). However, due to the inherent nature of causal inference methods relying on strong assumptions that are difficult to verify for observational health data, care must be taken when interpreting the causal task similarity measures.  

To this end, the study also analyzed feature importance estimates (Figure \ref{fig:feat_imp}) and patient characteristics (Figure \ref{fig:lr_odds_ratios}) associated with confident predictions of stroke. While many features and characteristics were related to known risk factors for stroke, others revealed instances of shortcut learning (spurious associations) \cite{brown2023detecting, zech2018variable} and a tendency for meta-learning to learn common patterns across tasks (e.g., age in Figure \ref{fig:lr_odds_ratios}). A post-training analysis of the MR meta-learning method suggests that task similarity-based meta-learning better reflects inter-task variability, with more variation in important features across tasks (Figures \ref{fig:feat_imp} and Table \ref{tab:feature_importance_correlation}). 

In conclusion, this study demonstrates that meta-learning with causal task similarity is an effective approach for improving generalizability of health prediction models and highlights considerations for applying this strategy. Future work could consider applying this to other data modalities, diseases, and generalizability scenarios (distribution shifts or deployment settings), building on current machine learning research trends of larger pre-trained models that need to be robustly adapted to specific applications (tasks) and populations of interest. 

\section{Software availability}

Code for implementing the methods and experiments presented in this work are publicly available through a GitHub repository: \href{https://github.com/sophiewharrie/meta-learning-hierarchical-model-similar-causal-mechanisms}{https://github.com/sophiewharrie/meta-learning-hierarchical-model-similar-causal-mechanisms}.

% ACKNOWLEDGEMENTS ONLY GO IN THE CAMERA-READY, NOT THE SUBMISSION
\section{Acknowledgements}

This work was supported by the Research Council of Finland (Flagship programme: Finnish Center for Artificial Intelligence FCAI, and grant numbers 358958 and 359567) and UKRI Turing AI World-Leading Researcher Fellowship, EP/W002973/1. We also acknowledge the computational resources provided by the Aalto Science-IT Project from Computer Science IT. This research has been conducted using data from UK Biobank (project ID: 77565) and FinnGen. 

We want to acknowledge the participants and investigators of the FinnGen study. The FinnGen project is funded by two grants from Business Finland (HUS 4685/31/2016 and UH 4386/31/2016) and the following industry partners: AbbVie Inc., AstraZeneca UK Ltd, Biogen MA Inc., Bristol Myers Squibb (and Celgene Corporation \& Celgene International II Sàrl), Genentech Inc., Merck Sharp \& Dohme LCC, Pfizer Inc., GlaxoSmithKline Intellectual Property Development Ltd., Sanofi US Services Inc., Maze Therapeutics Inc., Janssen Biotech Inc, Novartis AG, and Boehringer Ingelheim International GmbH. Following biobanks are acknowledged for delivering biobank samples to FinnGen: Auria Biobank (\url{www.auria.fi/biopankki}), THL Biobank (\url{www.thl.fi/biobank}), Helsinki Biobank (\url{www.helsinginbiopankki.fi}), Biobank Borealis of Northern Finland (\url{https://www.ppshp.fi/Tutkimus-ja-opetus/Biopankki/Pages/Biobank-Borealis-briefly-in-English.aspx}), Finnish Clinical Biobank Tampere (\url{www.tays.fi/en-US/Research\_and\_development/Finnish\_Clinical\_Biobank\_Tampere}), Biobank of Eastern Finland (\url{www.ita-suomenbiopankki.fi/en}), Central Finland Biobank (\url{www.ksshp.fi/fi-FI/Potilaalle/Biopankki}), Finnish Red Cross Blood Service Biobank (\url{www.veripalvelu.fi/verenluovutus/biopankkitoiminta}), Terveystalo Biobank (\url{www.terveystalo.com/fi/Yritystietoa/Terveystalo-Biopankki/Biopankki/}) and Arctic Biobank (\url{https://www.oulu.fi/en/university/faculties-and-units/faculty-medicine/northern-finland-birth-cohorts-and-arctic-biobank}). All Finnish Biobanks are members of BBMRI.fi infrastructure (\url{https://www.bbmri-eric.eu/national-nodes/finland/}). Finnish Biobank Cooperative -FINBB (\url{https://finbb.fi/}) is the coordinator of BBMRI-ERIC operations in Finland. The Finnish biobank data can be accessed through the Fingenious® services (\url{https://site.fingenious.fi/en/}) managed by FINBB.

Study subjects in FinnGen provided informed consent for biobank research, based on the Finnish Biobank Act. Alternatively, separate research cohorts, collected prior the Finnish Biobank Act came into effect (in September 2013) and start of FinnGen (August 2017), were collected based on study-specific consents and later transferred to the Finnish biobanks after approval by Fimea (Finnish Medicines Agency), the National Supervisory Authority for Welfare and Health. Recruitment protocols followed the biobank protocols approved by Fimea. The Coordinating Ethics Committee of the Hospital District of Helsinki and Uusimaa (HUS) statement number for the FinnGen study proposal is Nr F\_2020\_036.

\bibliographystyle{abbrvnat}
\bibliography{references.bib}    

\appendix

\section{Hyperparameters and other experiment settings}

Details of hyperparameters and other experiment settings used for the deep learning models are given in Table \ref{tab:hyperparameters}. Details for implementations of the task similarity measures are as follows. ICP: InvariantCausalPrediction R package\footnote{https://cran.r-project.org/web/packages/InvariantCausalPrediction/index.html}, using random allocation for experimental indices, with alpha set to 0.1, selection method as "boosting", maxNoVariables as 10, and maxNoVariablesSimult as 5. CHI2: scikit-learn package\footnote{https://scikit-learn.org/stable/modules/generated/sklearn.feature\_selection.chi2.html} with default parameters. MR: TwoSampleMR R package\footnote{https://mrcieu.github.io/TwoSampleMR/articles/introduction.html}, with instruments obtained through LD clumping (r2 $<$ 0.001, window size 10,000 kb) of GWAS summary statistics. DAG: modified version of the DiBS (Differentiable Bayesian Structure Learning) package \footnote{https://differentiable-bayesian-structure-learning.readthedocs.io/en/latest/index.html} implemented with JAX, using an Erdős-Rényi prior for the graph model, a linear Gaussian likelihood model, 200 steps for global model training, 20 steps for local model fine-tuning, and 1 particle for both global and local models. For all methods, task similarity was computed using cosine distance between the resulting task vectors.

\label{apx:expsettings}

\begin{landscape}
\begin{table*}[t]
\centering
\caption{Hyperparameter Search Space for Model Tuning Used in Experiments}
\label{tab:hyperparameters}
\begin{tabular}{llll}
\hline
\textbf{Hyperparameter} & \textbf{Distribution} & \textbf{Range/Values} & \textbf{Applicable Models} \\
\hline
nn\_prior\_name & Fixed & multivariate\_normal\_prior & All \\
hidden\_layer\_size\_longitudinal & q\_log\_uniform & [20, 160] & All \\
inner\_learning\_rate & log\_uniform & [1e-6, 1e-1] & All \\
inner\_temperature & log\_uniform & [1e-5, 1e-1] & All \\
global\_prior\_sigma & log\_uniform & [1e-5, 1] & All \\
model\_init\_log\_sds & uniform & [log(0.0001), 0] & All \\
num\_mc\_samples & Fixed & 10 & All \\
\hline
num\_inner\_updates & int\_uniform & [1, 9] & Hierarchical Models \\
outer\_temperature & log\_uniform & [1e-5, 1e-1] & Hierarchical Models \\
outer\_learning\_rate & log\_uniform & [1e-6, 1e-1] & Hierarchical Models \\
\hline
auxiliary\_learning\_rate & log\_uniform & [1e-6, 1e-1] & Hierarchical Models with Task Similarity \\
auxiliary\_temperature & log\_uniform & [1e-5, 1e-1] & Hierarchical Models with Task Similarity \\
num\_auxiliary\_updates & int\_uniform & [1, 9] & Hierarchical Models with Task Similarity \\
eta\_1 & uniform & [0, 5] & Hierarchical Models with Task Similarity \\
eta\_2 & uniform & [0, 5] & Hierarchical Models with Task Similarity \\
avg\_num\_active\_tasks & int\_uniform & [1, 6] & Hierarchical Models with Task Similarity \\
\hline
optimization\_metric & Fixed & AUROC & All (Bayesian Optimization settings) \\
perc\_test\_data & Fixed & 0.5 & All (Bayesian Optimization settings) \\
perc\_query\_data & Fixed & 0.5 & All (Bayesian Optimization settings) \\
k\_folds & Fixed & 2 & All (Bayesian Optimization settings) \\
minibatch\_size & Fixed & 5 & All (Bayesian Optimization settings) \\
\hline
\end{tabular}
\end{table*}
\end{landscape}

\section{Additional experiment results}

\begin{table}[ht]
\centering
\caption{Spearman Rank Correlation of Feature Importance Rankings}
\label{tab:feature_importance_correlation}
\resizebox{0.6\columnwidth}{!}{%
\begin{tabular}{|c|cccccc|}
\hline
\multicolumn{7}{|c|}{Meta-learning Baseline} \\
\hline
Task & I63 & I64 & G45 & I62 & I60 & I61 \\
\hline
I63 & 1.000 & 0.947 & 0.958 & 0.922 & 0.930 & 0.942 \\
I64 & 0.947 & 1.000 & 0.933 & 0.894 & 0.918 & 0.924 \\
G45 & 0.958 & 0.933 & 1.000 & 0.925 & 0.938 & 0.938 \\
I62 & 0.922 & 0.894 & 0.925 & 1.000 & 0.897 & 0.908 \\
I60 & 0.930 & 0.918 & 0.938 & 0.897 & 1.000 & 0.927 \\
I61 & 0.942 & 0.924 & 0.938 & 0.908 & 0.927 & 1.000 \\
\hline
\multicolumn{7}{|c|}{Task Similarity-based Meta-learning (MR)} \\
\hline
Task & I63 & I64 & G45 & I62 & I60 & I61 \\
\hline
I63 & 1.000 & 0.994 & 0.995 & 0.551 & 0.548 & 0.510 \\
I64 & 0.994 & 1.000 & 0.994 & 0.566 & 0.561 & 0.493 \\
G45 & 0.995 & 0.994 & 1.000 & 0.550 & 0.538 & 0.477 \\
I62 & 0.551 & 0.566 & 0.550 & 1.000 & 0.644 & 0.601 \\
I60 & 0.548 & 0.561 & 0.538 & 0.644 & 1.000 & 0.483 \\
I61 & 0.510 & 0.493 & 0.477 & 0.601 & 0.483 & 1.000 \\
\hline
\end{tabular}%
}
\end{table}

\begin{table}[ht]
\centering
\caption{Hyperparameter Values Selected by Bayesian Optimization for Different Task Similarity Measures. Values are presented as mean (standard deviation) across trials.}
\label{tab:bayesopresults}
\resizebox{0.6\columnwidth}{!}{%
\begin{tabular}{|c|c|c|c|}
\hline
\textbf{Method} & \textbf{avg\_num\_active\_tasks} & \textbf{eta\_1} & \textbf{eta\_2} \\
\hline
CHI2 & 5.00 (1.15) & 2.86 (1.73) & 2.49 (1.87) \\
ICP & 5.00 (0.82) & 1.82 (1.64) & 2.64 (1.79) \\
MR & 4.00 (1.83) & 0.98 (0.85) & 1.03 (0.75) \\
DAG & 5.75 (0.50) & 2.87 (1.11) & 3.41 (0.93) \\
\hline
\end{tabular}%
}
\end{table}

\end{document}